\documentclass{article}

\PassOptionsToPackage{numbers, compress}{natbib}

\usepackage{neurips_data_2024}
\usepackage{graphicx}
\usepackage{subfig}
\usepackage[utf8]{inputenc} 
\usepackage[T1]{fontenc}    
\usepackage{hyperref}       
\usepackage{url}            
\usepackage{booktabs}       
\usepackage{amsfonts}       
\usepackage{nicefrac}       
\usepackage{microtype}      
\usepackage{xcolor}         

\title{A Benchmark Time Series Dataset for Semiconductor Fabrication Manufacturing Constructed using Component-based Discrete-Event Simulation Models}

\author{%
  Vamsi Krishna Pendyala \\
  Arizona State University\\
  Tempe, AZ 85281 \\
  \texttt{vpendya2@asu.edu} \\
  \And
  Hessam S. Sarjoughian \\
  Arizona State University\\
  Tempe, AZ 85281 \\
  \texttt{hessam.sarjoughian@asu.edu} \\
  \AND
  Bala Sujith Potineni \\
  Arizona State University\\
  Tempe, AZ 85281 \\
  \texttt{bpotinen@asu.edu} \\
  \And
  Edward J. Yellig \\
  Intel Corporation\\
  Chandler, AZ 85226 \\
  \texttt{edward.j.yellig@intel.com} \\
}

\begin{document}

\maketitle

\begin{abstract}
Advancements in high-computing devices increase the necessity for improved and new understanding and development of smart manufacturing factories. Discrete-event models with simulators have been shown to be critical to architect, designing, building, and operating the manufacturing of semiconductor chips. The diffusion, implantation, and lithography machines have intricate processes due to their feedforward and feedback connectivity. The dataset collected from simulations of the factory models holds the promise of generating valuable machine-learning models. As surrogate data-based models, their executions are highly efficient compared to the physics-based counterpart models. For the development of surrogate models, it is beneficial to have publicly available benchmark simulation models that are grounded in factory models that have concise structures and accurate behaviors. Hence, in this research, a dataset is devised and constructed based on a benchmark model of an Intel semiconductor fabrication factory. The model is formalized using the Parallel Discrete-Event System Specification and executed using the DEVS-Suite simulator. The time series dataset is constructed using discrete-event time trajectories. This dataset is further analyzed and used to develop baseline univariate and multivariate machine learning models. The dataset can also be utilized in the machine learning community for behavioral analysis based on formalized and scalable component-based discrete-event models and simulations.
\end{abstract}

\section{Introduction}
\label{sec:Introduction}
Simulation of automated manufacturing processes is essential for the efficient use of complex and expensive machines. This requires effective coordinated decision-making due to a variety of factors including market conditions and processes spanning days to months with end-to-end controls of networked machines in minutes. Extensive resources are employed to create accurate physics-based manufacturing models to achieve these well-understood needs. In addition, designing, conducting, and evaluating simulation experiments are major undertakings. To reduce cost and time for practitioners, research in machine learning methods supporting building manufacturing digital twins is attracting greater emphasis among academic and industry researchers.
There are many benefits of machine learning models for analytics. Executing data-based models compared with their counterpart physics-based models is expected to be computationally scale-free. The concept of models primarily generated from data is aimed at responding faster to the changes that can lead to knowing how best manufacturing systems should operate under short/long time horizon requirements. Furthermore, generated ML model libraries can be frequently updated as additional data becomes available relative to physics-based models.
It is also possible to generate synthetic data using physics-based simulation of semiconductor factories. This class of data is strictly prescriptive and structured. The basic entities that undergo various kinds of processing include dies on wafers, wafers making lots, and wafer lots assembled into batches. Inventories, machines, and transportation are the entities that are used for processing semiconductor chips. Every entity can be ascribed one or more quantitative/qualitative variables. Every variable has a measurable value at any instance within a finite period. Other data are also measured and aggregated, such as average inventory volume, work-in-progress in machines, and wafers’ transportation routes. It is common for every machine to behave either deterministically or stochastically. The time designated to each entity in factories can be continuous or discrete. The collected data can have arbitrary accuracy and precision. Formal models of factories are suitable for creating sound datasets for use in developing regression and deep machine-learning models. Given the event-driven nature of manufacturing systems, Discrete-Event Simulation (DES) is widely used. This approach naturally defines inputs and outputs as events that can occur at any arbitrary time. The events can interrupt any process and the processes can be combined to form factories using well-formed input and output relationships.
However, the development of ML models depends on having rich data combined with expert domain knowledge among other factors. While massive amounts of live data are collected from semiconductor factories (e.g., from work centers to enterprise supply chain systems \citep{yuan2014}), it is challenging to generate ML models. In a semiconductor fabrication factory, data should be collected, organized, maintained, and synchronized across different individual and networked machines with varying logical/physical topologies in time and space. Given that collecting data from actual semiconductor factories is restricted due to the company’s proprietary constraints and the resources required for data engineering, benchmark datasets can be developed using simulations of physics-based models. Indeed, various efforts have been undertaken to collect data from physics-based models for use in statistical and machine-learning studies \citep{Korosteleva2021, Cachay2021}. 

\section{Background}
\label{sec:Background}
Manufacturing systems can be modeled as continuous differential equations, discrete-time, and discrete event specification languages \cite{zeigler2018theory}. Following system theory, they can be modeled as a set of atomic and coupled components. Every atomic model has the means to specify input, output, and state variables and state transition, output, and timing functions. Every coupled component has its own input and output variables. It has a set of atomic and other coupled models. Atomic and coupled models communicate through sending and receiving inputs (input variables) and outputs (output variables). Simulation output (data) naturally is partitioned and belongs to the atomic and coupled components. 
\subsection{Discrete-event modeling}
\label{sec:PDEVS model}
Discrete Event Simulation has been extensively used to model and simulate semiconductor manufacturing (e.g., \cite{monch2012production}). One of the methods for DES is known as Parallel Discrete Event System Specification (PDEVS) \cite{chow1994parallel}. An atomic model has distinct input, state, and output variables, each with a finite set of values. The variables can have primitive and compound alphanumeric values including null values. The input and output trajectories are event-based (i.e., every trajectory has only a finite number of values over a finite period). Every state trajectory is piecewise constant and has a value for every time instance. Every atomic model has external, internal, confluent, output, and timing functions. The external transition function is responsible for processing input events and changes of state. The internal transition function is responsible for state change in the absence of input events. The confluent function defined an order for the combined external and internal transition functions. Input events may received, and outputs may be sent at nonuniform time intervals. Multiple input and output events may occur concurrently at arbitrary time instances. State transitions can occur either due to receiving input events (external transition function) or not (internal transition function). Atomic and coupled models can be hierarchically composed to create other coupled models using external input, external output, and internal coupling relationships. Well-formed input and output for atomic and coupled components have concise physics-based syntax and semantics.

Parallel DEVS models are causal, providing concise understanding and rich interpretations of simulated behavior. Execution of these event-based atomic and coupled models results in time trajectories where their data do not necessarily have any uniform time intervals (i.e., for any of the input, output, and state variables may be separated at arbitrary time instances). These models can be simulated using a variety of simulators supported in popular programming languages and executable on single/multi-processor computing platforms supported with distributed technologies \cite{oren2023body}.  

\subsection{PDEVS Semiconductor Fabrication Model}
\label{sec:PDEVS Semiconductor Fabrication Model}
Single-stage and multi-stage semiconductor fabrication factory models are developed based on the PDEVS formalism \cite{chow1994parallel}. It is based on the description of a single-stage benchmark model named MiniFab \citep{sarjoughian2023}. The factory is modeled as a coupled model that has a Diffusion, Implantation, and Lithography machine. A machine is either in processing or repair mode at any given time. Every machine processes wafer lots in consecutive, non-interruptable loading, processing, unloading, and transportation phases, each with configurable duration and stochasticity. Each machine can enter a repair mode either after processing a number of lots or an amount of time representing the mean time between failures. A coordinator dispatches wafer lots to the diffusion machines. Similarly, another coordinator dispatches wafer lots to the Implantation machines. The assignment of wafer lots to these machines is instantaneous.

The factory models can receive Product a (Pa), Product b (Pb), and Test wafer (Tw) lots. The lots form a batch with a size of three before being chronologically processed in six steps starting from Diffusion to Implantation and ending in Lithography. Steps 1 and 5 are assigned to each of the diffusion machines named A and B, steps 2 and 4 are assigned to each of the implantation machines named C and D, and steps 3 and 6 are assigned to the lithography machine named E. Feedforward and feedback relationships among the machines define the ordering of the six steps (refer to Figure 1(a) for the diagram illustration of the single-stage factory \citep{sarjoughian2023}).

The number for the Pa and Pb can be {0, 2, 3, 6, 9, 12, 18, 24, 27, 30, 36, 45, 48, 54, 60, 72, 81, 90} and for Tw can be {0, 1, 3, 6, 9, 12, 15, 18, 27} \cite{sarjoughian2023}. Atomic models are developed to generate wafer lots in uniform. The Pa, Pb, and Tw wafer lots are generated every 8, 16, and 24 units of time (hours), respectively. Sinusoidal wafer lots are also generated at the same frequency, with sizes having simple patterns of 1, 2, 3, 2, 1. Transducers are devised to collect data from the machines and coordinators in each part of the factory. These do not influence the operations of the MiniFab. 
 
\section{Related work}
\label{sec:Related work}
Machine learning (ML) has been pivotal in advancing the traditional semiconductor manufacturing process \citep{irani1993applying, ailisto2023benefits}. As \citet{jiang2020novel} discusses, ML can be effectively utilized to improve the yield in semiconductor manufacturing. Various methods have been explored for data collection necessary for developing these ML models. For instance, \citet{liu2022machine} reviews the use of different ML algorithms and available datasets for enhancing semiconductor manufacturing. \citet{shin2000machine} derive data from manufacturing logs, while \citet{misc_eto} highlights a publicly available dataset of the semiconductor manufacturing process. The dataset \citep{misc_secom_179} offers time series data for a specific factory configuration, limiting its flexibility to alter factory settings and parameters.

In this research, we focus on using Discrete Event Simulation (DES) to generate data \citep{chan2022generation} for a given factory configuration, which can then be used to develop ML algorithms. Specifically, we aim to predict the overall throughput of the factory for a given time instance. Although \citet{godahewa2021monash} discusses various time series datasets and evaluation metrics along with the implementation of time series models, there is no exclusive dataset related to semiconductor manufacturing. The work by \citet{pr9030407} uses the MiniFab model description \cite{spier1995simulation} to develop a model in Anylogic. A dataset from this DES model is generated and used to develop ML classification models. We extend the functionalities of the simulation model to incorporate features such as repair state and wafer generation dataset. One key aspect of semiconductor manufacturing is predicting the throughput of a fabrication plant for a given time instance and factory configuration \citep{chong2016relationship}. We perform data analysis with different factory configurations and analyze parameters such as throughput and turnaround time of a semiconductor manufacturing process at different time instances.

\section{Dataset generation}
\label{sec:Dataset generation}
The PDEVS models are used to simulate a set of experiments conducted for the single-stage and multi-stage factories using the DEVS-Suite simulator \cite{DEVS-Suite7}. Four transducer models are used to measure and collect input, output, and state information at every simulation execution step. An eight-stage cascade model is created using the single-stage model. This model generates scenarios that exhibit higher structure and behavior complexities. Based on the logic of wafer processing of the MiniFab we formed 93 different tuples of Pa, Pb, and Tw values, each comprising different lot configurations ranging from small, medium, and large lot sizes relative to each other. In addition to the lot size and lot configurations, we can also specify how the model can process a particular batch of wafers (see Table~\ref{table:Simulation Scenarios}). We have 372 simulation scenarios for the eight-stage cascade factory models based on wafer lot sizes, configurations, uniform and sinusoidal patterns, and repair mode. Using the fabrication model, 93 experiments are simulated, each representing a scenario for Pa, Pb, and Tw. These simulated experiments are chosen to form small, medium, and large lot configurations with unique lot sizes. (see Table~\ref{table:Simulation Scenarios}). 

\begin{table}[h]
\centering
\begin{tabular}{cccc|c|c}
\hline
\multicolumn{4}{c|}{\textbf{Lot Configurations}}            & \textbf{}                 & \textbf{}                 \\ \hline
\textbf{Pa} & \textbf{Pb} & \textbf{Tw} & \textbf{Lot Size} & \textbf{Repair State}     & \textbf{Wafer Generation} \\ \hline
\multicolumn{4}{c|}{93 Lot Configurations}                  & Processing Steps          & Uniform                   \\
\multicolumn{4}{c|}{93 Lot Configurations}                  & Mean-time between failure & Uniform                   \\
\multicolumn{4}{c|}{93 Lot Configurations}                  & No Repair                 & Uniform                   \\
\multicolumn{4}{c|}{93 Lot Configurations}                  & No Repair                 & Sinusoidal                \\ \hline
\end{tabular}
\caption{Simulation Scenarios}
\label{table:Simulation Scenarios}
\end{table}

\begin{figure}[h]
    \centering
    \subfloat[]{\includegraphics[width=0.5\textwidth]{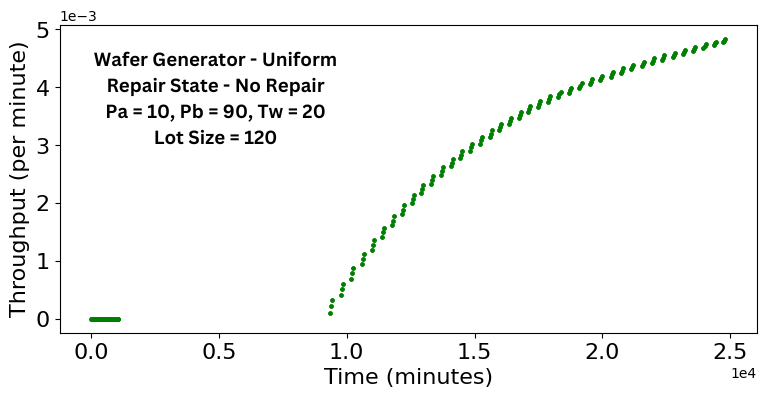}}
    \subfloat[]{\includegraphics[width=0.5\textwidth]{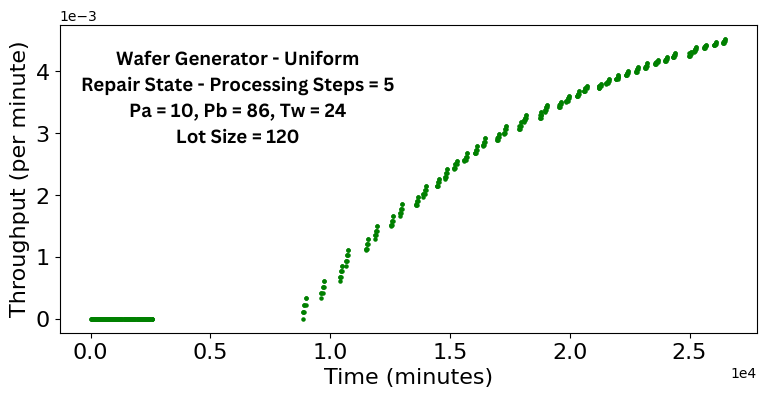}}
    \caption{MiniFab cascade factory simulation data}
    \label{fig:MiniFab cascade factory simulation data}
\end{figure}

The output generated from each simulation is stored in comma-separated value (CSV) files for individual atomic and coupled factory model components. Each simulated component has several rows, each representing an instance of time associated with the columns of data for every model (e.g., throughput shown in Figure~\ref{fig:MiniFab cascade factory simulation data}). For example, data on loading and transportation time trajectories are essential for determining optimal factory operation under different factory configurations. Data collected for throughput and turnaround time can be used to build surrogate machine-learning models, which can be trained to mimic the simulated behavior of the factory models as closely as possible.
Every of the simulation output files has a column depicting the logical time of the simulation, hence plotting different values against time can provide us with time series data. These values' temporal nature provides insight into wafer processing at each step. Our focus of this research is to analyze the throughput of a semiconductor manufacturing factory with different configurations. Now we plot the throughput values against time with an interval of 1 minute between every successive time value as depicted in Figure~\ref{fig:MiniFab cascade factory simulation data}. 

In these plots, we can see that there are some empty values at certain time instances, the empty values are based on the fact that our discrete event simulation records values at an event occurrence. The conversion of these values into a complete time series dataset would require us to have values at each time instance (Section~\ref{subsec:Time-series dataset}). Even though the plots have a similar monotonically increasing trend for different configurations the values at a granular level differ from each other. The differences can be visualized in the plots based on different configurations of the factory.

\subsection{Time series dataset}
\label{subsec:Time-series dataset}
\begin{figure}[h]
    \centering
    \subfloat[]{\includegraphics[width=0.5\textwidth]{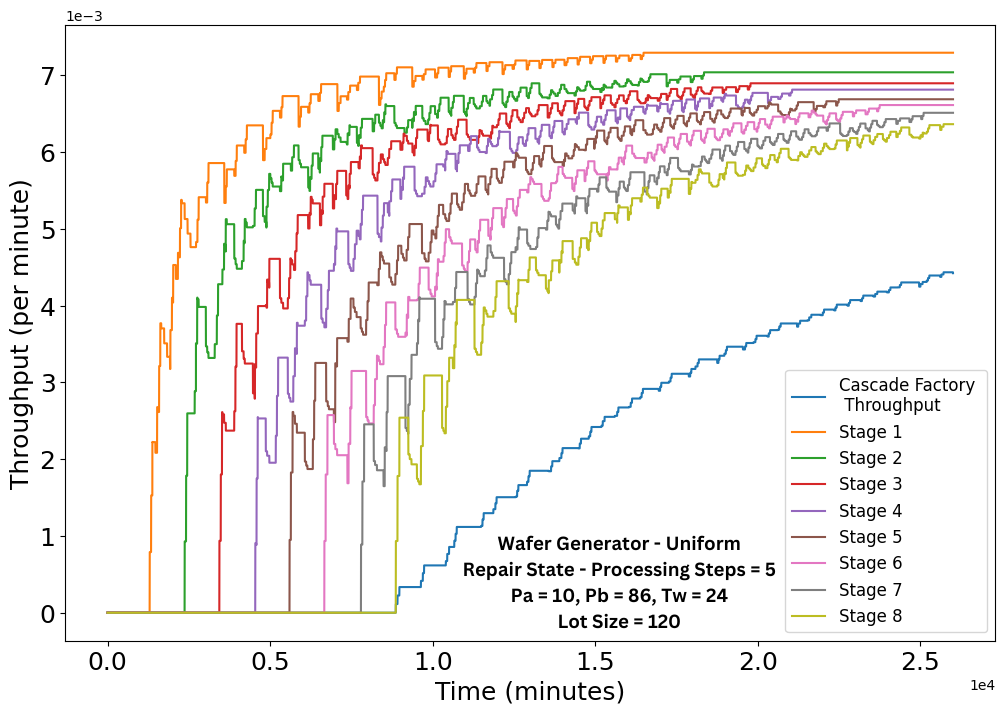}} 
    \subfloat[]{\includegraphics[width=0.5\textwidth]{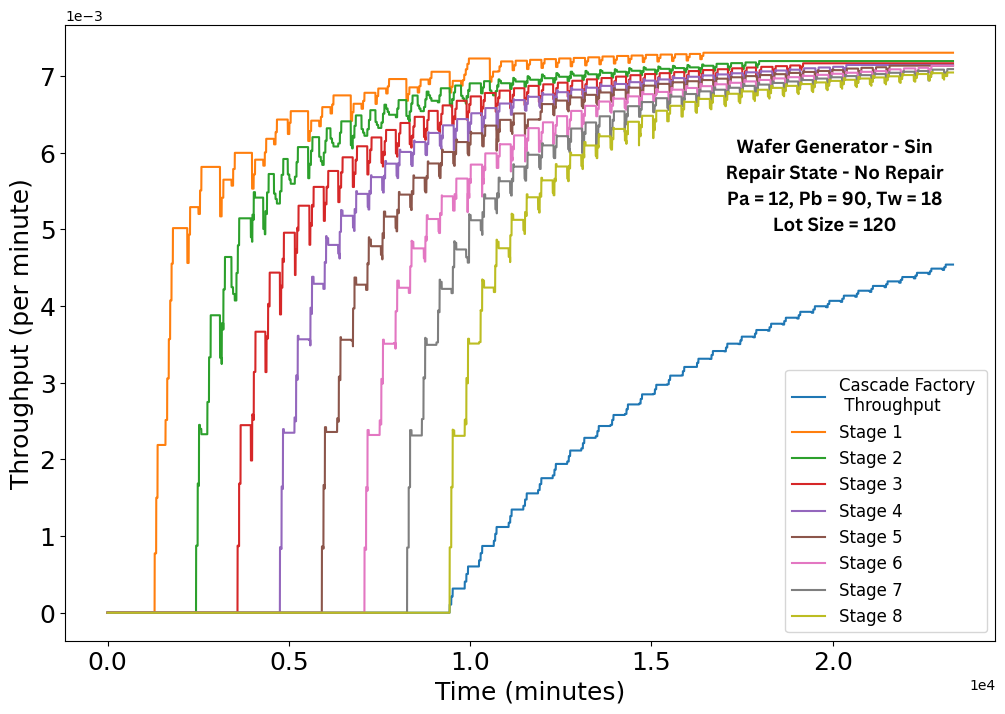}}
    \caption{MiniFab 8-stage cascade factory throughput time series}
    \label{fig:MiniFab 8-stage cascade factory throughput time-series}
\end{figure}

From the simulation output, we have data at different time instances of a MiniFab factory. Since values at certain time intervals are not directly provided by the simulation, for converting the data into a time series, we have used the front-filing method to fill the missing values. The reason for using a front-fill is based on the fact that the throughput values are piece-wise constant at different time instances until and unless they are changed by the occurrence of an event, hence the value at a time instance if not defined would be the value at a previous instance. This pre-processing has helped us to form a complete time series dataset with time granularity being 1 minute. Considering the dataset with a file for each stage's values, a univariate and a multivariate time series analysis can be performed depicting the versatility of the dataset. We can see the time series plots of multiple stages of a factory as shown in Figure~\ref{fig:MiniFab 8-stage cascade factory throughput time-series}. In these plots, we can see throughput values of each stage of an 8-stage cascade MiniFab factory for a given configuration, and the `Cascade Factory Throughput' in these plots indicates the overall throughput value of the factory. The plots indicate that introducing a repair state in the factory increases the level of uncertainty in the data. As per Section~\ref{sec:Background}, every stage of the MiniFab model is connected to the next stage (connected components), indicating a causality in throughput time series between each stage. Hence, can even perform a multivariate time series analysis to predict the throughput of each stage based on the throughput of other stages which is discussed further in Section~\ref{sec:Demonstration of benchmark datasets}. For a particular factory configuration, we can even configure the number of stages the factory can have, here we have considered an 8-stage cascade MiniFab factory.

\begin{figure}[h]
    \centering
    \subfloat[]{\includegraphics[width=0.5\textwidth]{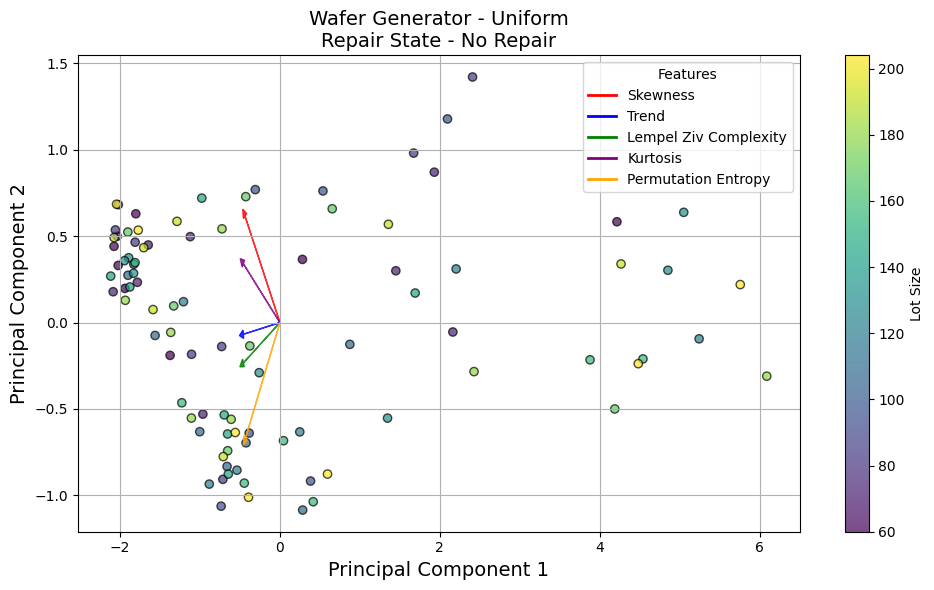}} 
    \subfloat[]{\includegraphics[width=0.5\textwidth]{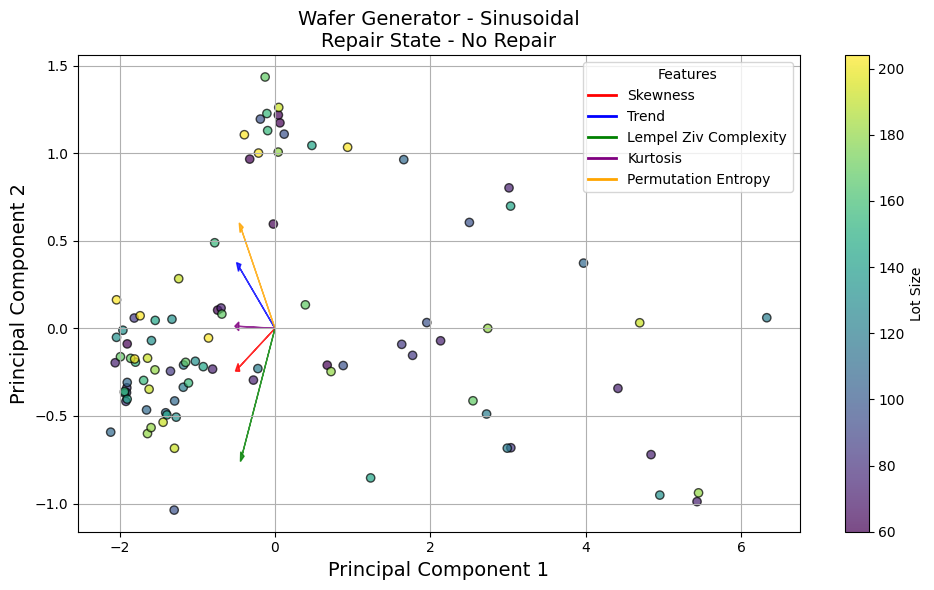}} 
    \hfill
    \subfloat[]{\includegraphics[width=0.5\textwidth]{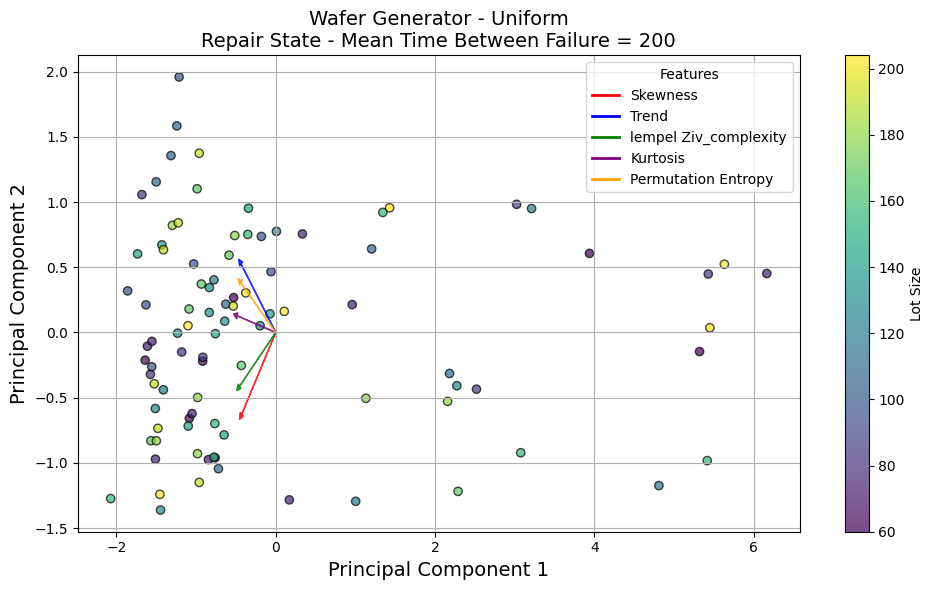}}
    \subfloat[]{\includegraphics[width=0.5\textwidth]{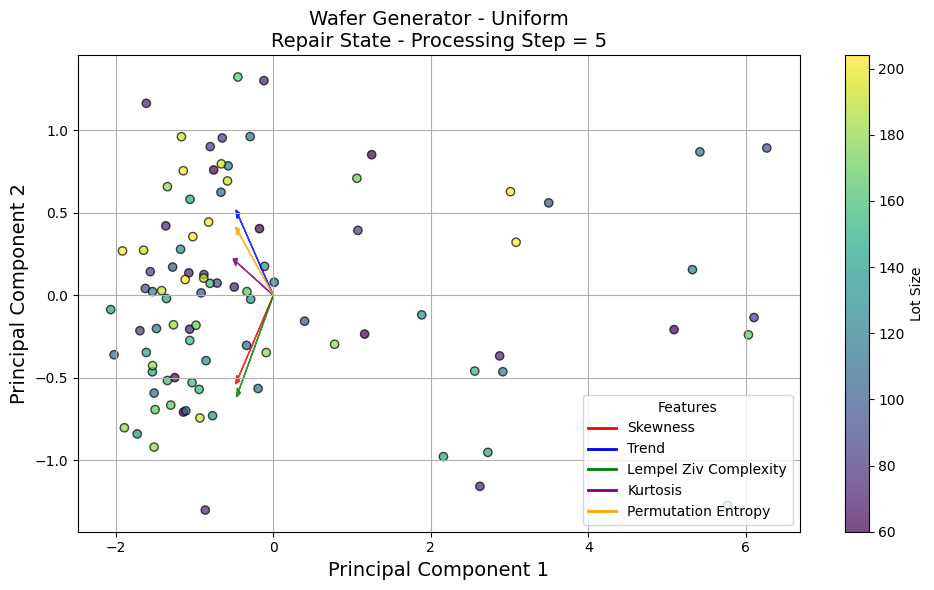}}
    \caption{PCA plots for throughput time series features}
    \label{fig:PCA plots for throughput time-series features}
\end{figure}

\subsection{Feature extraction and analysis}
\label{subsec:Feature extraction and analysis}
To better understand the relationship between throughput values of different configurations, we have extracted around 6,995 features of the 'Cascade Factory Throughput' time series for all the 372 simulation runs using Python's TsFresh library \cite{christ2018time}. These include the lag features (auto-correlation function, partial auto-correlation function, etc., for different lag values), trend, skewness, quantile changes, entropy, etc., for the time series. The values of all these features are present in GitHub\footnote{Repository: \url{https://github.com/comses/SCFM.git}}. The Principle Component Analysis depiction for the Linear Trend r-value, Skewness, Lempel Ziv Complexity \cite{ziv1977universal}, Kurtosis \cite{pearson1905criterion}, and Permutation Entropy are shown in Figure~\ref{fig:PCA plots for throughput time-series features}. Skewness provides valuable insights into the asymmetry of the data distribution, indicating potential irregularities or outliers in the manufacturing process. The linear trend r-values, which are close to 1 in our case for all the time series, signify a strong linear relationship over time, enabling predictive analysis of future outcomes and trends in the manufacturing process. We have observed negative kurtosis (K) values for all the time series, indicative of a platykurtic distribution, suggesting a more stable and uniform process with fewer extreme outliers, facilitating early detection of deviations from expected distributions. Lempel Ziv Complexity (LZC) measures the complexity of the manufacturing process, helping to identify patterns and irregularities that may impact production quality. Permutation entropy (PE) assesses the randomness and unpredictability within the data, offering insights into the information content and complexity of the manufacturing process. By leveraging these features collectively, semiconductor manufacturers can gain a comprehensive understanding of their production processes, identify areas for improvement, and optimize operational efficiency \cite{bojer2021kaggle}. The majority of the actual skewness values for all time series are positive and the actual linear trend r-values are closer to `1' due to the monotonically increasing nature of overall throughput with time (refer to Figure~\ref{fig:MiniFab 8-stage cascade factory throughput time-series}). The kurtosis values across all conditions are negative suggesting the series to be platykurtic with a lack of extreme values and a more uniform distribution, reflecting stability in the process. The features can be visualized using their PCA-Loadings plots as shown in Figure~\ref{fig:PCA plots for throughput time-series features}. In the scenario where the wafer generator is uniform and there is no repair state as depicted in Figure~\ref{fig:PCA plots for throughput time-series features} (a), data points exhibit clustering around the center. Skewness and trend emerge as dominant features influencing the variation in the data, suggesting a uniform process with some asymmetry. For factories with sinusoidal wafer generation as shown in Figure~\ref{fig:PCA plots for throughput time-series features} (b), fewer wafer lots are queued for processing resulting in a comparatively smaller skewness vector, indicating that generator configurations govern skewness. Also for the same scenario, data points are more dispersed, reflecting increased variability. Permutation entropy gains significance alongside skewness, indicating added complexity. In another scenario with a uniform generator and repair based on mean-time between failure in Figure~\ref{fig:PCA plots for throughput time-series features} (c), data points cluster closely, showing consistency. Skewness and trend remain significant, indicating a stable process with some asymmetry. Lastly, when the uniform generator is coupled with repair based on processing steps as in Figure~\ref{fig:PCA plots for throughput time-series features} (d), distinct clustering patterns emerge. The influence of complexity measures like Lempel Ziv Complexity and Permutation Entropy diminishes, reflecting reduced complexity. A comparison of these values based on the PCA-loading plots indicates that the factory with a uniform generator with no repair has a lesser overall throughput. Additionally, by adding a repair state, we have more cluster points near the trend vector indicating a higher throughput. 

\section{Demonstration of benchmark datasets}
\label{sec:Demonstration of benchmark datasets}

\begin{figure}[h]
    \subfloat[]{\includegraphics[width=0.50\textwidth]{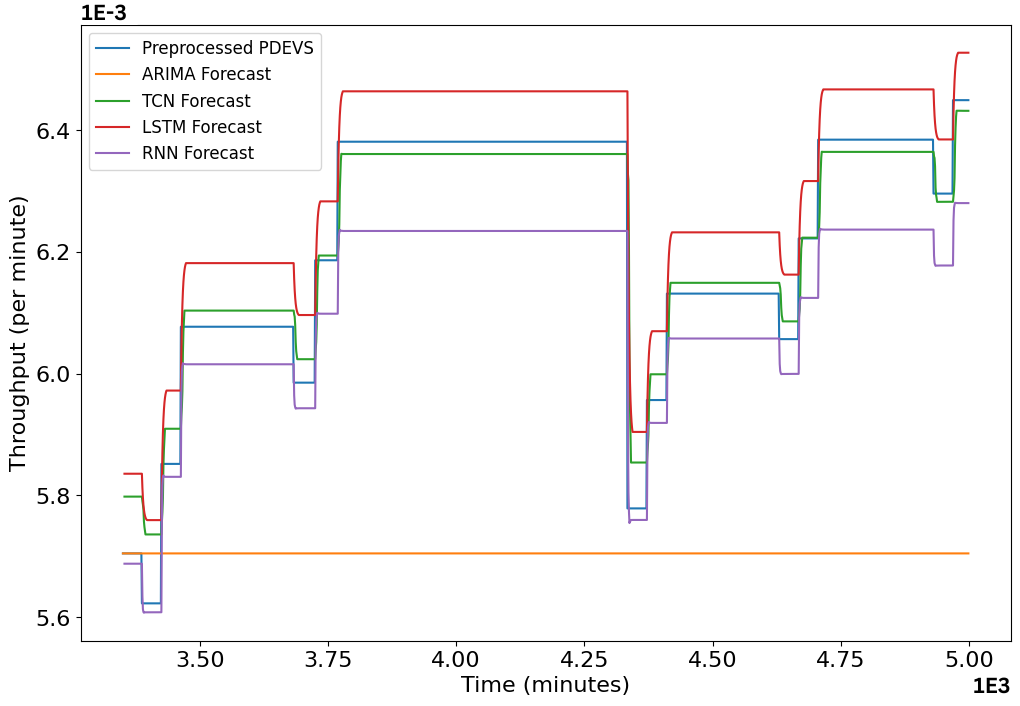}}
    \subfloat[]{\includegraphics[width=0.47\textwidth]{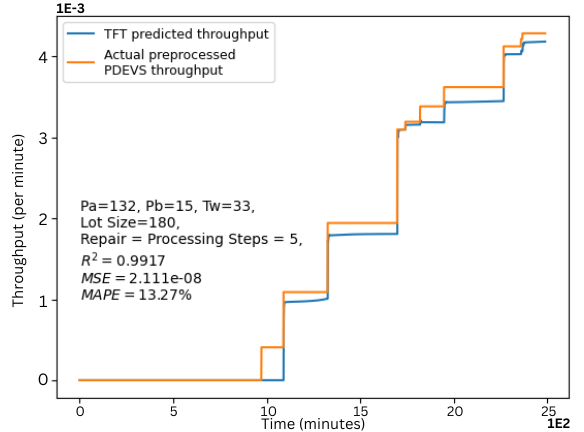}}
    \caption{Comparison of PDEVS with ARIMA, RNN, LSTM, TCN and TFT}
    \label{fig:Throughput TA Comparison}
\end{figure}

The simulation output is transformed from a discrete event (Figure~\ref{fig:MiniFab cascade factory simulation data}) to a discrete-time time series (Figure~\ref{fig:MiniFab 8-stage cascade factory throughput time-series}) by filling the missing values with previous values. For constructing a baseline model, we have used multiple time series forecasting models: Auto-Regressive Integrated Moving Average (ARIMA), Recurrent Neural Network (RNN), Long-Short Term Memory (LSTM), Temporal Convolutional Neural Network (TCN) and Temporal Fusion Transformers (TFT). The choice of selecting these models was based on the uniqueness of the dataset considering the large size of the dataset (25,000 instances) and the monotonically increasing nature with granular differences in the values of throughput for different configurations. ARIMA captures linear trends and seasonality \cite{box1970time}, RNN and LSTM handle sequential data and dependencies \cite{elman1990finding, hochreiter1997long}, TCN models long-range temporal dependencies \cite{bai2018empirical} and TFT model considers the impact of static variables (e.g., repair state, lot size, etc.) for time series forecasting \cite{lim2021temporal}. As per feature analysis in Section~\ref{subsec:Feature extraction and analysis}, some models may outperform others. Choosing the right metrics to evaluate the performance of these models on the dataset is significant. Mean-square error (MSE), is significant for evaluating forecasting performance, but since, Mean-Square Error is not a scale-free error metric and due to the small scale ($\sim10^{-3}$) of throughput values, we get small scale values ($\sim10^{-6}$) of MSE, hence it is convenient to use MSE as a relative measure to evaluate a model's performance. The value of Mean Average Percentage Error (MAPE) can be used to assess the performance of a model as it is scale-free, but since MAPE cannot handle cases where the actual value of throughput can be zero, we consider non-zero values of throughput for computing MAPE. Next, we used $R^2$ scores to evaluate the performance, which is both scale-independent and can handle zero values in the dataset, but it can be misleading for non-linear relationships, sensitive to outliers, and does not account for overfitting or model complexity. Therefore, $R^2$ should be used with other metrics for a comprehensive evaluation of model performance. Finally, we use the Mean-Forecast Error (MFE) which directly compares the actual and predicted values hence it is scale-dependent, for throughput prediction it can be of the same scale as the actual values ($\sim10^{-3}$). All the above metrics can be used as a combination to evaluate the performance of a model, since the dataset consists of multiple zero values and a small scale, relying on a single error metric can lead to ambiguous results. Additionally, with these metrics, it is also pertinent to visualize the prediction plots as there are granular differences in the overall throughput values. Performance evaluation of different time series models for an exemplary simulation configuration: Pa = 10, Pb = 90, Tw = 20, Lot Size = 120, Repair State = No Repair, Wafer Generator = Uniform results are as depicted in Figure~\ref{fig:Throughput TA Comparison} (a). Since the TFT model considers also the static covariates of a time series, the evaluation for the TFT model has been done for different configurations as seen in Figure~\ref{fig:Throughput TA Comparison} (b). The time series models predict throughput values based on a look-back window of size 10 and the error metrics for the model evaluation are as mentioned in Table~\ref{table:timeseries model comparison}.

\begin{table}[]
\center
\begin{tabular}{@{}llllll@{}}
                        & \multicolumn{5}{c}{\textbf{Model}}                                          \\ \midrule
\textbf{Error Metric} & \textbf{ARIMA} & \textbf{RNN} & \textbf{LSTM} & \textbf{TCN} & \textbf{TFT} \\ \midrule
MSE                     & 3E-07          & 1.35E-08     & 9.73E-09      & 2.29E-09     & 2.11E-08     \\
R2   Score              & -6.006         & 0.681        & 0.771         & 0.946        & 0.9917       \\
MFE                     & 5.07E-04       & 1.04E-04     & -9.35E-05     & -4.56E-06    & -9.59E-03    \\
MAPE                    & 0.0813         & 0.0167       & 0.0153        & 0.0051       & 0.1327       \\ \bottomrule
\end{tabular}
\caption{8-stage MiniFab time series model comparison}
\label{table:timeseries model comparison}
\end{table}

The ARIMA model has an MSE value of $3\times10^{-7}$ and a MAPE value of $8\%$ but the prediction plot of the ARIMA model shows sub-optimal performance by the model. Hence the error metric along with the actual plot serves as a strong parameter when compared to error metrics alone in the case of throughput prediction.

\subsection{Univariate time series forecasting}

\begin{figure}[h]
    \centering
    \subfloat[Throughput prediction $R^2$ scores]{\includegraphics[width=0.51\textwidth]{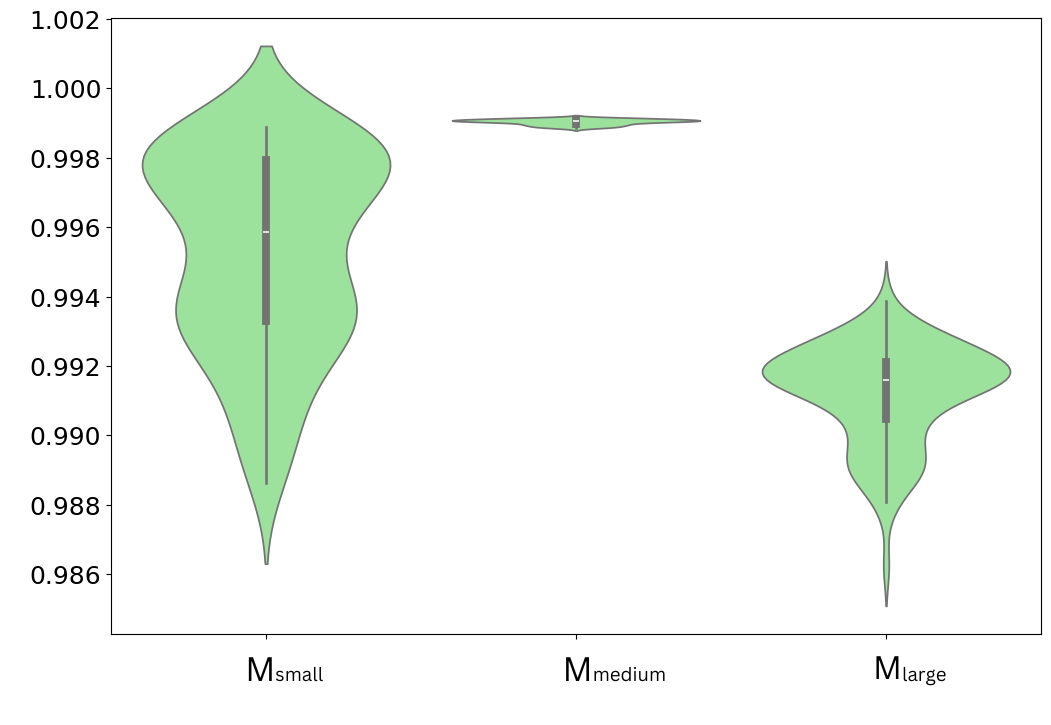}}
    \subfloat[Turnaround time prediction $R^2$ scores]{\includegraphics[width=0.49\textwidth]{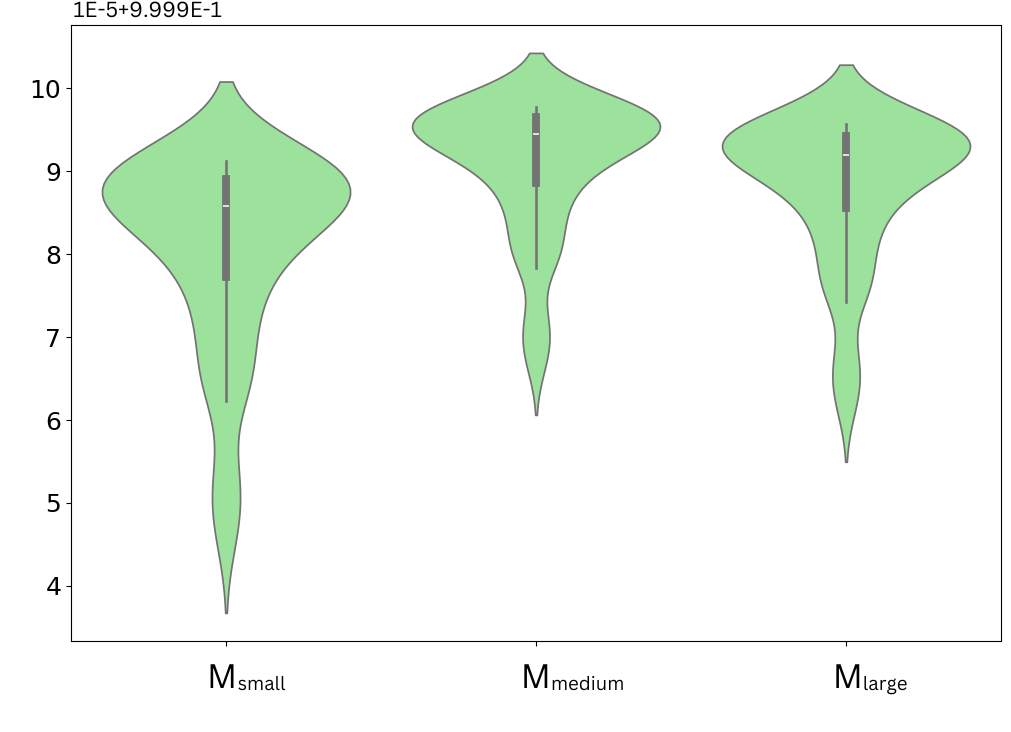}}
    \caption{Cascade factory lot configuration and size impact \cite{pendyala2024}}
    \label{fig:Cascade factory lot size impact}
\end{figure}

Based on the PCA-Loading plots in Figure~\ref{fig:PCA plots for throughput time-series features}, the data points are spread for different lot sizes based on lot configuration, this suggests that lot size doesn't share a linear relationship with time series forecasting. With our dataset derived from various lot configurations (Section~\ref{sec:Dataset generation}), we aim to identify an optimal lot size for training a time series model to enhance accuracy and understand the impact of lot size on model performance. We employed a TCN model, training it on three lot size categories relatively: small - $M_{small}$(60: Pa=15, Pb=36, Tw=9), medium - $M_{medium}$ (120: Pa=10, Pb=90, Tw=20), and large - $M_{large}$ (192: Pa=150, Pb=24, Tw=18) with a uniform wafer generator and no repair. Comprehensive tests across all 93 configurations were followed to provide insights into achieving high accuracy in time series models based on optimal lot sizes. The prediction $R^2$ scores are depicted in the Figure~\ref{fig:Cascade factory lot size impact} for models $M_{small}$, $M_{medium}$, and $M_{large}$. The results corroborate our observations in Section~\ref{subsec:Feature extraction and analysis} as the model trained on medium lot size ($M_{medium}$) has a better performance compared to other models, additional results of turnaround time prediction also show similar behavior as seen in Figure~\ref{fig:Cascade factory lot size impact} (b).

\subsection{Multivariate time series forecasting}

\begin{figure}[h]
    \centering
    \subfloat[$\mathcal{TH}$ prediction for $\mathcal{S}_2$ based on $\mathcal{S}_1$ \& $\mathcal{S}_2$]{\includegraphics[width=0.48\textwidth]{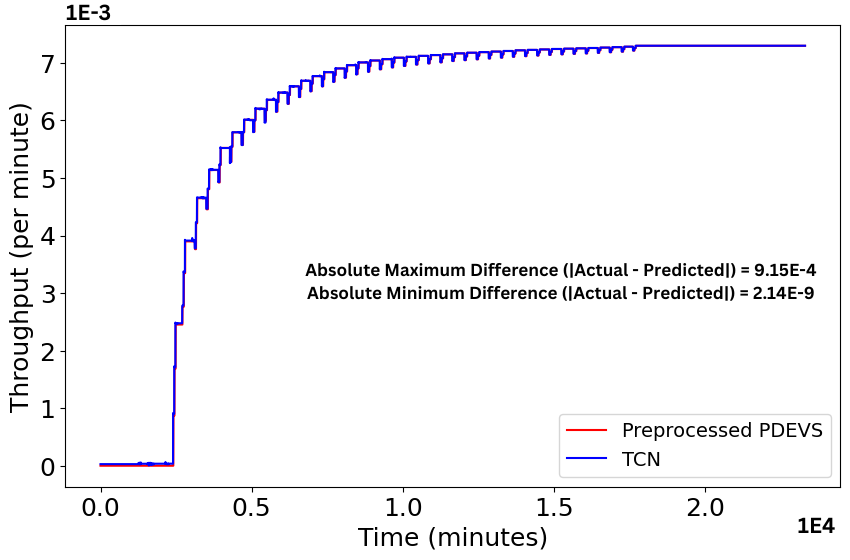}}
    \subfloat[$\mathcal{TH}$ prediction for $\mathcal{S}_5$ based on $\mathcal{S}_3$ \& $\mathcal{S}_4$]{\includegraphics[width=0.48\textwidth]{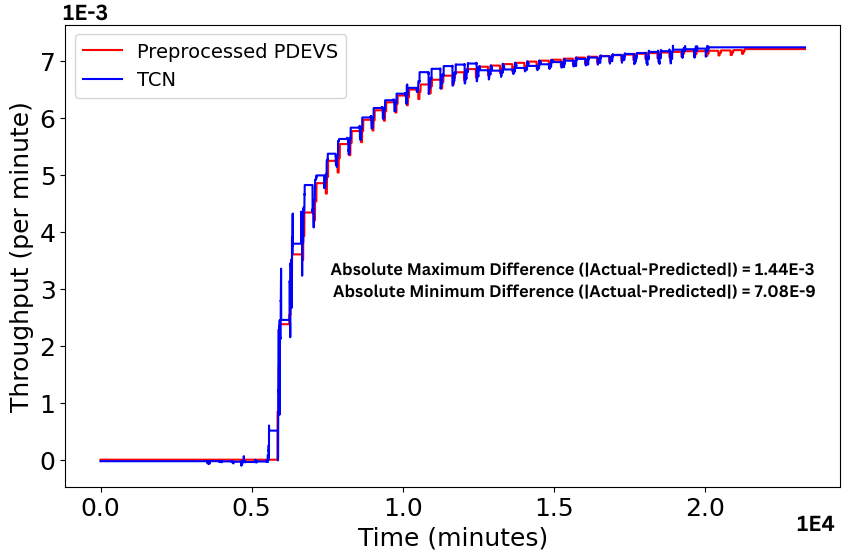}}
    \caption{Multivariate prediction \cite{pendyala2024}}
    \label{fig:Multivariate Prediction Best-Worst}
\end{figure}

Section~\ref{sec:Background} suggests that each stage of the MiniFab model is a connected component to other stages. This suggests that a multivariate analysis can be performed between the throughput values of each stage. The plots in Figure~\ref{fig:MiniFab 8-stage cascade factory throughput time-series} show how each stage throughput values for a similar trend. We performed multivariate analysis using the TCN model to predict overall or selected stage throughput, considering various combinations of throughput values from multiple stages as input. The TCN model demonstrated strong performance in multivariate analysis, as shown in Figure~\ref{fig:Multivariate Prediction Best-Worst}. These findings emphasize the importance of considering stage interdependence in semiconductor manufacturing predictive modeling, with higher accuracy noted towards the simulation's end and with more stages.

\section{Conclusion}

Machine learning (ML) models suitable for time series can benefit from concise and accurate datasets collected from physics-based simulations. These ML models are desirable due to their significant execution speedup compared to simulating causal models. The ML datasets facilitate a wide range of metrics and measurements. Principal Component Analysis (PCA) captures key time series features and helps understand the role and impact of various semiconductor fabrication factory configurations on essential operational measures. We have developed datasets using PDEVS models and Intel’s benchmark factory description. We detailed dataset generation for multi-stage factories for uniform (with and without repair/maintenance) and sinusoidal wafer lot experimental configurations. The utility of these datasets is demonstrated by developing time series models, such as TCN and TFT. They can make predictions comparable to those obtained from PDEVS simulations, with TFT accounting for static covariates. Future work includes developing additional models and benchmark datasets for semiconductor supply-chain systems.

\begin{ack}
This research is funded by Intel Corporation, Chandler, Arizona, USA.
\end{ack}

\bibliographystyle{plainnat}
\bibliography{references}

\end{document}